\documentclass[runningheads]{llncs}
\usepackage{comment}
\usepackage{soul,color}
\usepackage{epsfig}
\usepackage{epstopdf}
\usepackage{graphicx}
\usepackage{amsmath}
\usepackage{amssymb}
\usepackage{multirow}
\usepackage{rotating}
\usepackage{subcaption}
\usepackage{booktabs}
\usepackage{algorithm}
\usepackage{algpseudocode}

\begin{document}
\pagestyle{headings}
\mainmatter
\def\ECCVSubNumber{5389}  

\title{ShuffleBlock: Shuffle to Regularize Deep Convolutional Neural Networks} 

\titlerunning{ShuffleBlock}
%
\author{Sudhakar Kumawat \and
Gagan Kanojia \and
Shanmuganathan Raman}
\authorrunning{S. Kumawat et al.}
%
\institute{Indian Institute of Technology Gandhinagar, Gandhinagar, India, 382355}
\maketitle

\begin{abstract}
    Deep neural networks have enormous representational power which leads them to    overfit on most datasets. Thus, regularizing them is important in order to reduce overfitting and enhance their generalization capabilities. Recently, channel shuffle operation has been introduced for mixing channels in group convolutions in resource efficient networks  in order to reduce memory and computations. This paper studies the operation of channel shuffle  as a regularization technique in deep convolutional networks. We show that while random shuffling of channels during training drastically reduce their performance, however, randomly shuffling small patches between channels significantly improves their performance. The patches to be shuffled are picked from the same spatial locations in the feature maps such that a patch, when transferred from one channel to another, acts as structured noise for the later channel. We call this method ``ShuffleBlock''. The proposed ShuffleBlock module is easy to implement and improves the performance of several baseline networks  on the task of image classification on CIFAR and ImageNet datasets. It also achieves comparable and in many cases better performance than many other regularization methods. We provide several ablation studies on selecting various hyperparameters of the ShuffleBlock module and propose a new scheduling method that further enhances its performance.
\keywords{Convolutional neural networks, Regularization, Channel shuffle}
\end{abstract}

\section{Introduction}\label{sect:intro}


\begin{figure}[t]
	\begin{center}
		\includegraphics[width=0.5\linewidth]{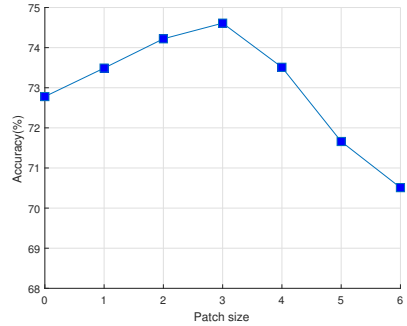}
	\end{center}
	\caption{\textbf{Test accuracy v/s size of the patch size}. The variation of the test accuracies obtained on CIFAR-100 using ResNet-110 with ShuffleBlock while varying the value of patch size. Patch size $=0$ means no ShuffleBlock applied. Patch size equal to 3 performs best.}
	\label{fig:ablation_patchsize}
\end{figure}
\begin{figure}[t]
	\begin{center}
		\includegraphics[width=\linewidth]{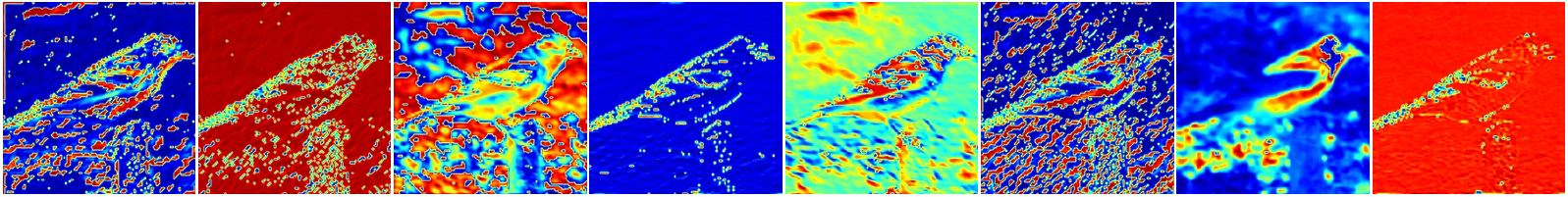}
		\includegraphics[width=\linewidth]{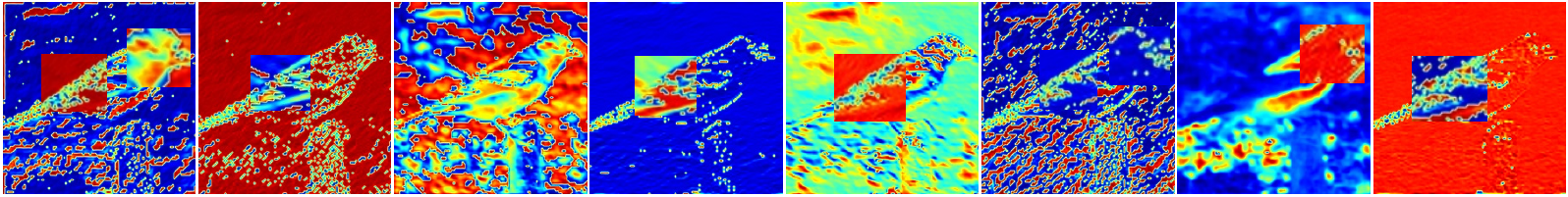}
	\end{center}
	\caption{\textbf{Visualization of feature maps before and after applying ShuffleBlock}. Network used is ResNet-50. Row-1  shows the activation maps output by the first layer of the ResNet-50 network with no regularization (excepth Batch Normalization). Row-2  presents the output when ShuffleBlock is applied to these activation maps.}
	\label{fig:intro_2}
\end{figure}
In recent years, with the availability of enormous computational power and large datasets, research in the area of deep neural nets such as convolutional neural networks (CNNs) has led to unprecedented advances in various computer vision problems such as image classification \cite{he2016deep,hu2018squeeze,szegedy2016rethinking}, video classification \cite{hara2018can,tran2015learning}, object detection \cite{liu2016ssd,ren2015faster}, and semantic segmentation \cite{chen2018deeplab,long2015fully}. Deep CNNs have enormous representational power. However, with this boon comes the curse of overfitting which prevents them from reaching their optimum performance levels. \\
In order to curb overfitting and improve performance, various regularization and attention techniques have been proposed. In model regularization space, methods such as Dropout \cite{srivastava2014dropout,choe2019attention}, DropBlock \cite{ghiasi2018dropblock}, and StochDepth \cite{huang2016deep} prevent CNNs from focusing too much on a small set of intermediate activations by randomly dropping features maps or regions in intermediate feature maps thus improving the generalization and discriminative power of the CNN model. Other related methods such as Cutout \cite{devries2017improved} and Random erase augmentation \cite{zhong2017random} apply the above ideas on the input image itself. Such methods are called data augmentation techniques.\\
The above methods mainly rely on removing regions in feature maps/images while training in order to introduce noise for regularization. In this work, we study channel shuffle operation as a means to generate noise for regularization. Recently, the channel shuffle operation has been shown to be effective in CNNs for combining features in group convolutions and for reducing computations significantly \cite{zhang2017interleaved,zhang2018shufflenet}. In this work, we study the channel shuffle operation as a general method for regularizing any network. We show that shuffling whole channels  randomly during training leads to a drastic drop in performance. However,  randomly  shuffling  small  patches between channels significantly improves their performance if the patches to be shuffled are picked from the same spatial locations in the feature maps. Fig.~\ref{fig:ablation_patchsize} shows a glimpse of our study on the test accuracy versus shuffled patch size for the ResNet-110 network on the CIFAR-100 dataset. Here, patch size corresponding to the value zero means that no shuffling of channels was done. We observe that if the patch size to be shuffled between channels are too small ($1\times 1$) or very large (whole channel), the network does not generalize better. However, it performs better when it is in the range between 2 and 4. Note that the patch size may depend on the spatial resolution of the input image. Furthermore, the patches to be shuffled are picked from the same spatial locations in the channels such that a patch when transferred form one channel to another acts as structured noise for the later channel as shown in the third row of Fig.~\ref{fig:intro_2}. We call the above method of inter channel patch shuffling as ``ShuffleBlock''. In \cite{kang2017patchshuffle}, the author shuffles the pixels within the each local patch of randomly selected feature maps which is different from our approach.\\
Fig.~\ref{fig:intro_2} gives a visual understanding of the proposed ShuffleBlock method. Here, the  first row shows the activation maps output by the first layer of the ResNet-50 network trained on the ImageNet dataset with no regularization (except Batch Normalization) applied. Second row presents the output when ShuffleBlock  is applied to these activation maps. Note that the  ShuffleBlock method,  randomly shuffles contiguous regions of feature maps between channels.\\
In summary, the main contributions of this work are as follows.
\begin{itemize}
	\item We explore the channel shuffle operation from the perspective of network regularization and propose a new method called ShuffleBlock.  
	\item We show that ShuffleBlock improves the performance of several baseline networks  on the task of image classification on CIFAR and ImageNet datasets.
	\item  We also propose a new scheduling method called ``Inverted Step Scheduling'' that when used with ShuffleBlock during training further improves its performance. 
	\item We provide several ablation studies on selecting various hyperparameters of the ShuffleBlock module.
\end{itemize}

\section{Related Works}
Due to their large representational capacity, CNNs are  prone to overfitting. In order to counter overfitting, various regularization techniques have been proposed. These techniques can  be broadly categorised into data  and functional spaces.\\

\noindent\textbf{Regularization Techniques in Function Space.} These techniques include methods such as Dropout \cite{srivastava2014dropout}, DropBlock \cite{ghiasi2018dropblock} and StochDepth \cite{huang2016deep} that regularize the internal features of the CNN models. Dropout \cite{srivastava2014dropout} regularizes deep neural nets by randomly dropping nodes in the fully connected layers during training time. DropBlock \cite{ghiasi2018dropblock} generalizes Dropout to convolutional layers by dropping contiguous regions in feature maps while training. StochDepth \cite{huang2016deep} randomly drops a subset of layers and bypasses them with identity functions during the training of very deep neural networks, thus reducing training time and improving accuracy significantly. The proposed ShuffleBlock technique regularizes the network in the function space of the model. Here the removed regions are filled with patches  from other feature maps as noise. Other methods that regularize neural nets in functional space include weight decay \cite{krogh1992simple} and Batch Normalization \cite{ioffe2015batch}.\\

\noindent\textbf{Regularization Techniques in Data Space.} These techniques include  data augmentation and synthesizing methods such as Cutout \cite{devries2017improved}, Label smoothing \cite{szegedy2016rethinking}, and Mixup \cite{zhang2017mixup}. Cutout \cite{devries2017improved} consists of masking out random sections of input images during training by simulating occluded examples and encouraging the model to take more minor features into consideration when making decisions, rather than relying on the presence of a few major features. Label smoothing \cite{szegedy2016rethinking} is a regularization technique for classification problems. It selectively manipulates labels to prevent the model from predicting the training examples too confidently.  Mixup \cite{zhang2017mixup} synthesizes new training examples through weighted linear interpolation of two existing examples, thus increasing and stabilizing the performance of models especially generative adversarial networks.\\
An important drawback of the regularization techniques in data space compared to those in functional space is that their application to structured prediction problems such as image segmentation  is not obvious.
\begin{figure}[t]
	\begin{center}
		\includegraphics[width=0.6\linewidth]{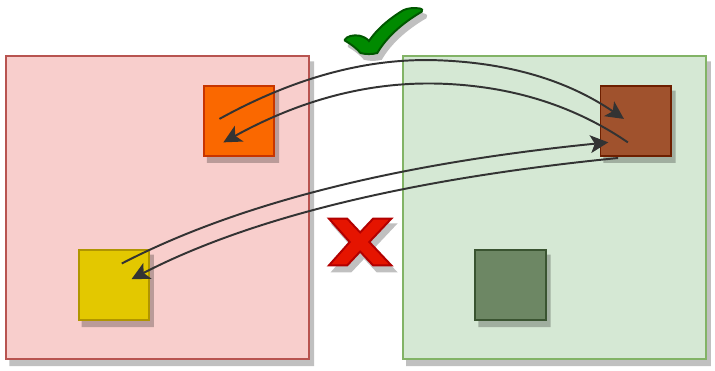}
	\end{center}
	\caption{\textbf{ShuffleBlock: Selecting patches among channels for shuffling}.  The patches to be shuffled are picked from the same spatial locations in the feature maps such that a patch when transferred from one channel to another acts as structured noise for the channels.}
	\label{fig:long}
\end{figure}
\begin{figure}[t]
	\begin{center}
		\includegraphics[width=\linewidth]{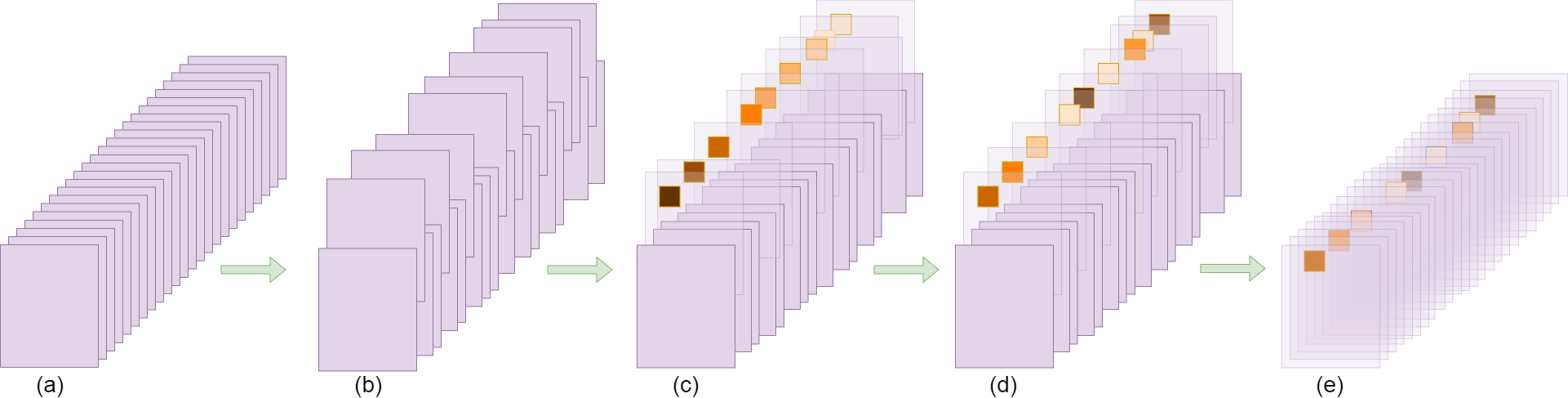}
	\end{center}
	\caption{\textbf{Visualization: Steps of ShuffleBlock}. (a) Input activation maps from a previous layer. (b) Randomly selecting $ch\_frac$ channels from the activation maps. (c) Randomly selecting spatial locations on the selected $ch\_frac$ channels and extracting patches of size $block\_size$. (d-e) Shuffling the patches among the channels.}
	\label{fig:shuffle_demo}
\end{figure}
\section{Method}
Our proposed ShuffleBlock is a simple yet effective method for regularizing CNNs with no overhead during inference and no addition of trainable parameters. It uses inter-channel shuffling of contiguous regions/patches in order to achieve higher performance. This can be visualized as introducing structured noise into feature maps where the noise is just a region/patch from another feature map. We call it structured noise because while shuffling patches between two feature maps, it is made sure that the patches are removed from the same location on the feature maps as shown in Fig.~\ref{fig:long}.
\subsection{ShuffleBlock}
Consider an intermediate feature map of size $C\times H\times W$ where $C$, $H$, and $W$, denote the number of channels, height, and width, respectively of the feature map. The ShuffleBlock has two main parameters $ch\_frac$ and $block\_size$. The parameter $ch\_frac$ controls the fraction of channels $C$ among which the patches will be shuffled. The parameter $block\_size$ controls the size of the patch/region in the feature map that will be shuffled.\\
First, ShuffleBlock takes an intermediate feature map of size  $C\times H\times W$ as input and randomly (uniform) selects $ch\_frac*C$ channels from it where each channel is of size $H\times W$. Next, it randomly (uniform) samples a 2d coordinate $i,j$ ($0\leq i\leq H$, $0\leq j\leq W$) and extracts square patches of size $block\_size$ with $i,j$ as the top-left corner from the selected channels. Note that the position $i,j$ is common for all the selected channels. Finally, it randomly (uniform) shuffles these patches among the selected channels. Note that the remaining unselected $C*(1-ch\_frac)$ channels are not affected by the above process. In  Fig.\ref{fig:shuffle_demo} we provide a visualization of the method discussed above. The pseudocode for ShuffleBlock is presented in Algorithm~\ref{alg: ShuffleBlock}. Furthermore, similar to Dropout \cite{srivastava2014dropout}, we do not apply ShuffleBlock during inference.\\

\begin{algorithm}[t]
	\begin{algorithmic}[1]
		\State \textbf{Input}:  Activation maps $A$ from previous layer with size $C\times H\times W$ , $block\_size$, $ch\_frac$, mode
		\If{mode $==$ Inference}
		\State return $A$
		\EndIf
		\State Randomly sample $ch\_frac * C$ number of channels from $A$.
		\State Randomly sample patches $M:M_{i,j}$ of size $block\_size$ from the above sampled channels. 
		\State Randomly shuffle the $M_{i,j}$ patches among the sampled channels.
	\end{algorithmic}
	\caption{ShuffleBlock}
	\label{alg: ShuffleBlock}
\end{algorithm}
\noindent\textbf{Setting the value of} \emph{block\_size}. In our experiments, we use constant $block\_size$ for all the feature maps, regardless of the resolution of feature map. We experimented with various values of $block\_size$ and found that patch sizes in the range of $3\times 3$ and $5\times 5$ give better performance. Details of these experiments are provided in Section~\ref{sec:exp}.\\

\noindent\textbf{Setting the value of} \emph{ch\_frac}.  In our experiments, we start  shuffling patches among $0.5-0.6$ fraction of the channels at the start of the training and and then decrease this fraction to $0$
towards the end of the training in a step-wise manner with the drop in learning rate. This scheduling method produces optimal performance for ShuffleBlock and is discussed in next paragraph. \\

\noindent\textbf{Inverted Step Scheduling.}  We observed that ShuffleBlock with fixed $ch\_frac$ during training achieves better accuracy. However,  decreasing  it during training further improves the performance. Furthermore, we observed that instead of gradually decreasing $ch\_frac$,  ShuffleBlock performs better if $ch\_frac$ is decreased step-wise with a step-wise drop in learning rate. In our experiments, we decrease $ch\_frac$ and learning rates in the same step. Note that the above scheduling method is technically opposite to the scheduling methods used in feature drop methods such as DropBlock \cite{ghiasi2018dropblock}. In DropBlock, the network starts its training with no drop in feature map regions for the first few epochs meaning the network runs as the baseline network (no DropBlock) for the first few epochs. Thereafter, the regions are dropped in a linearly increasing fashion with training thus improving the robustness and performance of the model. In ShuffleBlock, since $ch\_frac$ is decreased in a step-wise fashion, the network runs as the baseline network (no ShuffleBlock) in the final epochs during training.  In order to distinguish our proposed scheduling method with the above method, we call our method as ``Inverted Step Scheduling''.
\section{Experiments and Discussion}\label{sec:exp}
\subsection{Network Architectures}
\begin{figure}[t]
	\begin{center}
		\includegraphics[width=\linewidth]{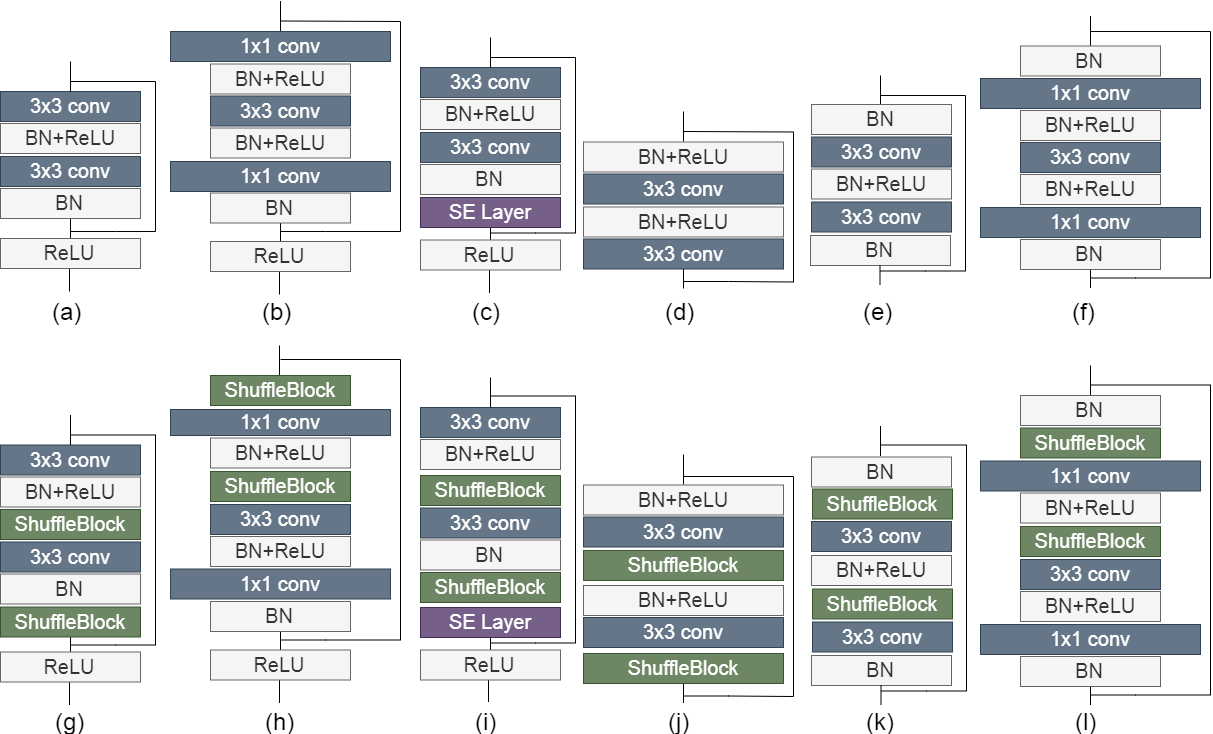}
	\end{center}
	\caption{(a) and (b) show the basic and bottleneck residual block used in ResNet (56 and 110 layers) and ResNet (50 layers), respectively. (c) and (d) show the residual blocks used in SENet (56 and 110 layers) and Wide Residual Network, respectively. (e) and (f) show the basic and bottleneck pyramidal residual blocks used in PyramidNet-48 and PyramidNet-200, respectively. (g), (h), (i), (j), (k), and (l) show the residual blocks with proposed ShuffleBlock for the residual blocks in (a), (b), (c), (d), (e), and (f), respectively.}
	\label{fig:shuffle_blocks}
\end{figure}

To show the effectiveness of ShuffleBlock, we use residual networks as the back-bone architectures. We employ ResNet (50, 56 and 110 layers) \cite{he2016deep}, squeeze-and-excitation networks (56 and 110 layers) \cite{hu2018squeeze}, wide residual networks (16 and 28 layers) \cite{zagoruyko2016wide} and PyramidNet (110 and 200 layers) \cite{han2017deep} for the experiments. These residual networks are the combination of residual building blocks connected in series. Fig. \ref{fig:shuffle_blocks} (a) and (b) show the basic and bottleneck residual blocks used in ResNet (56 and 110 layers) and ResNet (50 layers), respectively. Fig. \ref{fig:shuffle_blocks} (c) and (d) show the residual blocks used in SENet (56 and 110 layers) and Wide Residual Networks, respectively. Fig. \ref{fig:shuffle_blocks} (e) and (f) show the basic and bottleneck pyramidal residual blocks used in PyramidNet-48 and PyramidNet-200, respectively. For the experiments, we incorporate the ShuffleBlock in these residual networks as shown in Fig.(g), (h), (i), (j), (k), and (l). It can be observed that we have inserted the proposed ShuffleBlock without altering the overall structure of the residual blocks. This is done to evaluate the effectiveness of the proposed block when it is included in any base model.
\subsection{ImageNet Classification}
\noindent The ILSVRC 2012 classification dataset \cite{deng2009imagenet} comprises 1.2 million images for training, 50K images for validation and 150K images for testing from 1000 classes. For data augmentation, we adopt standard setting for ImageNet dataset such as resizing, random cropping, and flipping, as done in \cite{yun2019cutmix}. For evaluation, we applied the single-crop scheme \cite{zagoruyko2016wide}. As per the common practice, we report classification accuracy on the validation set.
\begin{table}[t]
	\centering
	\begin{tabular}{lccc}  
		\toprule
		&&\multicolumn{2}{c}{\textbf{ImageNet}} \\
		\cmidrule(r){3-4}
		\textbf{Model} & \textbf{\#Params}  &\multicolumn{1}{c}{\textbf{Top 1}} & \multicolumn{1}{c}{\textbf{Top 5}} \\
		\midrule
		
		ResNet-50 \cite{he2016deep} & 25.56M & 75.30 &  92.20 \\
		ResNet-50 + BAM  \cite{park2018bam} & 25.92M & 75.98 &  92.82 \\
		ResNet-50 + Cutout \cite{devries2017improved} & 25.56M &  76.52 &  93.21 \\
		ResNet-50 + SE  \cite{hu2018squeeze} & 28.09M &  76.71 &  93.38 \\
		ResNet-50 + dropout (kp=0.7) \cite{srivastava2014dropout} & 25.56M &  76.80 &  93.41 \\
		ResNet-50 + DropPath (kp=0.9) \cite{larsson2016fractalnet} & 25.56M &  77.10 &  93.50 \\
		ResNet-50 + label smoothing (0.1) & 25.56M &  77.17 &  93.45 \\
		ResNet-50 + GALA  \cite{linsley2018global} & NA &  77.27 &  93.65 \\
		ResNet-50 + CBAM  \cite{woo2018cbam} & 28.09M &  77.34 &  93.69 \\
		ResNet-50 + ShuffleB + IS & 25.56M &  \textbf{77.42} &  \textbf{93.7}\\
		\bottomrule
	\end{tabular}
	\vspace{0.5em}
	\caption{\textbf{Performance results on the ImageNet dataset}. The comparison of ShuffleBlock with various methods in feature and data spaces on the ResNet-50 network. $kp$ stands for $keep\_prob$.\vspace{-2em}}
	\label{tab:imagenet}
\end{table}
\subsubsection{Experiments and Results}
Table \ref{tab:imagenet} shows the comparison of the validation accuracy obtained on ImageNet using the proposed approach with the state-of-the-art approaches. It can be seen that the proposed approach outperforms the baseline method with a significant margin. Also, among the state-of-the-art regularization techniques, the proposed approach performance is only next to DropBlock \cite{ghiasi2018dropblock}. Please note that the difference between the validation accuracy of DropBlock and ShuffleBlock  can be attributed to the fact that the hyperparameters of DropBlock are specifically fine-tuned for the ImageNet dataset. However, in our experiment for the ImageNet dataset, we use the hyperparameters that we fine-tuned on the CIFAR datasets.
We train ResNet-50 on ImageNet using stochastic gradient descent (SGD) using a batch size of 256 images.  We train the network for 300 epochs. The initial learning rate is set to 0.1 which is further reduced by a factor 10 on $75^{th}, 150^{th}$, and $225^{th}$ epoch. We train the network with inverted step scheduling. We keep the intial value of $ch\_frac$ equal to 0.5, and then, drop the value of $ch\_frac$ to 0.4, 0.3 and 0 at epoch 150, 225, and 275, respectively. We use a weight decay of $10^{-4}$.  We train the network using the standard data augmentation with scale jittering as done in \cite{han2017deep}. We train the network using 4 GPUs. Originally, ResNet does not use dropout in the architecture. For dropout baseline, it is applied on the convolution branches. For DropPath baseline, DropPath is applied on all connections except the skip connections.
\subsection{CIFAR Classification}
\subsubsection{Dataset}
The CIFAR-10 and CIFAR-100 datasets are colored natural image datasets \cite{krizhevsky2009learning}. Both consists of $32\times 32$ pixel images. CIFAR-10 consists of images belonging to 10 classes with 6K images per class. CIFAR-100 consists of images belonging to 100 classes with 600 images per class. The training and test sets of both the datasets comprise 50K and 10K images, respectively. For data augmentation, we adopt the standard scheme, i.e., random cropping and flipping, for both the datasets  \cite{lin2013network}. 
\begin{table}[t]
	\centering
	\begin{tabular}{lcccccccc}  
		\toprule
		& & &\multicolumn{3}{c}{\textbf{CIFAR 10}} \\
		\cmidrule(r){4-6} 
		\textbf{Method} & \textbf{Depth} & \textbf{\#Params} & \textbf{Baseline} &  \textbf{ShuffleB} & \textbf{ShuffleB + IS}\\
		\midrule
		ResNet-56 \cite{he2016deep}&	56	& 0.85M &	93.03	& 94.08 &  94.69 \\
		ResNet-110 \cite{he2016deep} & 110 & 1.73M &	93.57 & 94.69 & 94.90 \\
		SENet-56 \cite{hu2018squeeze} &	56 & 0.86M & 94.07 & 94.42 & 94.73\\
		SENet-110 \cite{hu2018squeeze} &	110	& 1.74M &	94.79	& 95.14 & 95.47\\
		WRN-16x8 \cite{zagoruyko2016wide} & 16 &	11M & 95.19 & 95.71 & 95.43 \\
		WRN-28x10 \cite{zagoruyko2016wide} &	28 & 36.5M	& 95.83 & 96.36 & 96.22\\
		PyramidNet-48 \cite{han2017deep}	& 110	& 1.77M	& 95.42	& 95.87 & 95.79\\
		\bottomrule
	\end{tabular}
	\vspace{0.5em}
	\caption{\textbf{Performance results on the CIFAR-10 image classification datasets}. The comparison of the ShuffleBlock based networks with their corresponding baselines. Here, ShuffleB and IS refer to ShuffleBlock and Inverted Step Scheduling,  respectively.}
	\label{tab:baseline_results1}
\end{table}
\begin{table}[t]
	\centering
	\begin{tabular}{lcccccccc}  
		\toprule
		& & &\multicolumn{3}{c}{\textbf{CIFAR 100}}\\
		\cmidrule(r){4-6} 
		\textbf{Method} & \textbf{Depth} & \textbf{\#Params} & \textbf{Baseline} &  \textbf{ShuffleB} & \textbf{ShuffleB + IS}  \\
		\midrule
		ResNet-56 \cite{he2016deep}&	56	& 0.85M &  72.33 & 73.6  & 74.50    \\
		ResNet-110 \cite{he2016deep} & 110 & 1.73M &	72.78 &	75.08 & 76.14 \\
		SENet-56 \cite{hu2018squeeze} &	56 & 0.86M & 73.85 &	75.15 &  75.62 \\
		SENet-110 \cite{hu2018squeeze} &	110	& 1.74M & 74.91 & 76.55 & 76.47 \\
		WRN-16x8 \cite{zagoruyko2016wide} & 16 &	11M & 77.93 &	79.27 & 78.98 \\
		WRN-28x10 \cite{zagoruyko2016wide} &	28 & 36.5M &	79.5 & 80.89 & 80.83 \\
		PyramidNet-48 \cite{han2017deep}	& 110	& 1.77M	& 76.87 & 76.99 & 78.55 \\
		\bottomrule
	\end{tabular}
	\vspace{0.5em}
	\caption{\textbf{Performance results on the CIFAR-100 image classification datasets}. The comparison of the ShuffleBlock based networks with their corresponding baselines. Here, ShuffleB and IS refer to ShuffleBlock and Inverted Step Scheduling,  respectively. }
	\label{tab:baseline_results2}
\end{table}
\begin{table}[t]
	\centering
	\begin{tabular}{lcc}  
		\toprule
		&\multicolumn{2}{c}{\textbf{CIFAR 100}} \\
		\cmidrule(r){2-3}
		\textbf{Model}  &\multicolumn{1}{c}{\textbf{Top 1}} & \multicolumn{1}{c}{\textbf{Top 5}} \\
		\midrule
		PyramidNet-200  \cite{han2017deep}    &  83.55 & 96.31    \\
		PyramidNet-200 + StochDepth \cite{huang2016deep} &  84.14 & 96.67 \\
		PyramidNet-200 + DropBlock \cite{ghiasi2018dropblock} & 84.27 & \textbf{96.74} \\
		PyramidNet-200 + ShuffleB + IS & \textbf{84.53} & 96.58\\
		\bottomrule
	\end{tabular}
	\vspace{0.5em}
	\caption{\textbf{Performance results on the CIFAR-100 dataset}. The comparison of ShuffleBlock with feature removal methods on the PyramidNet-200 network. In case of DropBlock, $keep\_prob$ is equal to 0.9 \cite{ghiasi2018dropblock}.}
	\label{tab:pyrm200_functional}
\end{table}
\begin{table}[t]
	\centering
	\begin{tabular}{lcc}  
		\toprule
		&\multicolumn{2}{c}{\textbf{CIFAR 100}} \\
		\cmidrule(r){2-3}
		\textbf{Model}  &\multicolumn{1}{c}{\textbf{Top 1}} & \multicolumn{1}{c}{\textbf{Top 5}} \\
		\midrule
		PyramidNet-200  \cite{han2017deep}    &  83.55 & 96.31      \\
		PyramidNet-200 + Cutout \cite{devries2017improved} & 83.47 & 96.35 \\
		PyramidNet-200 + Manifold Mixup \cite{verma2019manifold} & 83.86 & 95.93 \\
		PyramidNet-200 + Mixup \cite{zhang2017mixup} & 84.37 & 96.01\\
		PyramidNet-200 + ShuffleB + IS & \textbf{84.53} & \textbf{96.58} \\
		\bottomrule
	\end{tabular}
	\vspace{0.5em}
	\caption{\textbf{Performance results on the CIFAR-100 dataset}. The comparison of ShuffleBlock with methods in data space on the PyramidNet-200 network. In case of Manifold Mixup and Mixup, $\alpha$ is equal to 1 \cite{verma2019manifold,zhang2017mixup}.}
	\label{tab:pyrm200_data}
\end{table}
\subsubsection{Comparisons with the baselines}
\label{sec:Comparisons with the baselines}
To show the effectiveness of the proposed approach, we incorporate it in the state-of-the-art networks. We experimented with ResNet (56 and 110 layers), squeeze-and-excitation networks (56 and 110 layers), wide residual networks (16 and 28 layers) and PyramidNet-48 (110 layers). Table \ref{tab:baseline_results1} and \ref{tab:baseline_results2} show the comparison of the accuracy obtained using the proposed approach, i.e., networks with ShuffleBlock, with the baseline networks on CIFAR-10 and CIFAR-100. We experimented with two variations of the proposed approach: random shuffling and random shuffling with inverted step scheduling. It can be observed that the proposed method outperforms the baseline methods with significant margin. \\
We train the networks on CIFAR-10 and CIFAR-100 using stochastic gradient descent (SGD).  For these experiments, we train using a batch size of 64 images. We train ResNet, squeeze-and-excitation networks, and wide residual networks for 400 epochs. The initial learning rate is set to 0.1 which is further reduced by a factor of 10 on $160^{th}, 240^{th}, 320^{th}$, and $360^{th}$ epoch. When training the networks without inverted step scheduling, we keep $ch\_frac$ equal to 0.5 for the whole duration. When training the networks with inverted step scheduling, we keep the intial value of $ch\_frac$ equal to 0.6, and then, drop the value of $ch\_frac$ to 0.5, 0.4 and 0 at epoch 160, 240, and 360, respectively.\\
In case of PyramidNet-48, we train the network for 300 epochs while keeping the initial learning rate set to 0.1 which is further reduced by a factor of 10 on $150^{th}, 225^{th}$, and $275^{th}$ epoch. When training the networks without inverted step scheduling, we keep $ch\_frac$ equal to 0.5 for the whole duration.  When training with inverted step scheduling, we keep the initial value of $ch\_frac$ equal to 0.6, and then, drop the value of $ch\_frac$ to 0.5, 0.4 and 0 at epoch 150, 225, and 275, respectively. We use a weight decay of $10^{-4}$. We train all the networks on a single GPU except PyramidNet-48 which has been trained with 2 GPUs.
\subsubsection{Comparisons with other methods}
Table \ref{tab:pyrm200_functional} and \ref{tab:pyrm200_data} show the comparison of the proposed approach against the regularization and data augmentation methods on CIFAR-100, respectively. We experimented with PyramidNet-200 to show the comparison of the proposed method with regularization techniques in the functional space as well in the data space. The proposed approach lies among regularization techniques in the functional space like StochDepth \cite{huang2016deep} and DropBlock \cite{ghiasi2018dropblock}. It can be seen that the proposed approach outperforms the state-of-the-art regularization techniques in the functional space and the data space.\\
we train the network for 300 epochs while keeping the initial learning rate set to 0.25 which is further reduced by a factor of 10 on $150^{th}, 225^{th}$, and $275^{th}$ epoch. We train using a batch size of 64 images. We train the network with inverted step scheduling while keeping the intial value of $ch\_frac$ equal to 0.5, and then, drop the value of $ch\_frac$ to 0.4, 0.3 and 0 at epoch 150, 225, and 275, respectively. We use a weight decay of $10^{-4}$. We train the network using 2 GPUs.
\section{Ablation Studies}
\begin{table}[t]
	\centering
	\begin{tabular}{lcc}  
		\toprule
		\textbf{$block\_size$}  & CIFAR-10 & CIFAR-100 \\
		\midrule
		0 (baseline)  & 93.57  &  72.78   \\
		1  & 94.61  &  73.49   \\
		2  & 94.61  &   74.22  \\
		3  &  \textbf{94.69}  &  \textbf{74.61}   \\
		4  &  94.66 &   73.51  \\
		5  &  93.78 &  71.66   \\
		6  & 92.47  &  70.51   \\
		channel size & 25.48  &  7.00   \\
		channel size (reverse)  & 91.86  &  67.42   \\
		\bottomrule
	\end{tabular}
	\vspace{0.5em}
	\caption{\textbf{Exploring $block\_size$ for shuffling among channels on the CIFAR datasets}. The comparison of the test accuracies obtained on CIFAR datasets using ResNet-110 with ShuffleBlock while varying the value of block\_size. Patch/block size in the range 2 to 4 performs the best.}
	\label{tab:ablation_patchsize}
\end{table}
\subsection{Effect of Patch Size}
In this study, we observe the effect of size of the patches being randomly shuffled among the channels. We experimented with ResNet (110 layers) on CIFAR-10 and CIFAR-100. It can be observed in Fig. \ref{tab:ablation_patchsize} that initially from patchsize 1 to 3, the validation accuracy rises and later, its starts dropping gradually. Table \ref{tab:ablation_patchsize} shows comparison of the test accuracies obtained on CIFAR datasets using ResNet-110 with ShuffleBlock while varying the value of patch size. It can also be noticed that when we randomly shuffle the selected channels instead of patches, i.e.,the size of the patch is equal to the channel size, the network hardly learns anything (second last row of Table \ref{tab:ablation_patchsize}). We also experimented with the case when instead of randomly shuffling the selected channels, we reverse the order of the selected channels randomly (last row of Table \ref{tab:ablation_patchsize}). In this case, the network performs better than the case of random shuffling the selected channels. However, it still under performs than the case when the patches are randomly shuffled among the channels.\\
For the experiments, we train the networks on CIFAR-10 and CIFAR-100 using stochastic gradient descent (SGD). We train ResNet (110 layers) for 400 epochs. The initial learning rate is set to 0.1 which is further reduced by a factor of 10 on $160^{th}, 240^{th}, 320^{th}$, and $360^{th}$ epoch. For these experiments, we train using a batch size of 128 images. We train the networks without inverted step scheduling while keeping $ch\_frac$ equal to 0.5 for the whole duration. We use a weight decay of $10^{-4}$. We train these networks on a single GPU.
\begin{table}[t]
	\centering
	\begin{tabular}{lcc}  
		\toprule
		$ch\_frac$ & CIFAR-10 & CIFAR-100 \\
		\midrule
		0 (baseline)  & 93.57  &  72.78  \\
		0.1  & 93.58  &  72.80  \\
		0.2  & 94.58  &   74.04  \\
		0.4  &  94.41 &  \textbf{74.49}   \\
		0.5 & \textbf{94.68}  &  74.11   \\
		0.6  & 94.27  &  74.31  \\
		0.8 & 93.93  &  73.73  \\
		1.0  & 94.38  &  73.64  \\
		\bottomrule
	\end{tabular}
	\vspace{0.5em}
	\caption{\textbf{Exploring $ch\_frac$ for shuffling among channels on the CIFAR datasets}.The comparison of the test accuracies obtained on CIFAR datasets using ResNet-110 with ShuffleBlock while varying the value of $ch\_frac$. $ch\_frac$ in the range between $0.4$ and $0.6$ performs the best.}
	\label{tab:ablation_ch_frac}
\end{table}
\subsection{Inverted Step Scheduling}
In this study, we observe the effect of varying the fraction of patches being randomly shuffled among the channels during the training. We experimented with ResNet (110 layers) on CIFAR-10 and CIFAR-100. In inverted step scheduling, we reduce the $ch\_frac$ at certain epochs and towards the end of the training we stop shuffling the patches and train the network for few epochs without shuffling. Table \ref{tab:baseline_results1} and \ref{tab:baseline_results2} show the comparison between the results obtained with and without inverted step scheduling. The training details are provided in Section \ref{sec:Comparisons with the baselines}
\subsection{Effect of $ch\_frac$}
In this study, we observe the effect of the fraction of patches being shuffled among the channels. We experimented with ResNet (110 layers) on CIFAR-10 and CIFAR-100. We vary the values of $ch\_frac$ from 0 to 1. Table \ref{tab:ablation_ch_frac} shows the comparisons of the test accuracy obtained on CIFAR-10 and CIFAR-100 with different values of $ch\_frac$. It can be observed from Table \ref{tab:ablation_ch_frac} that the network performs well when $ch\_frac$ is 0.4-0.6. We train these networks without inverted step scheduling while keeping patch size equal to 3. Other hyperparameter settings are same as the ablation study for the effect of patch size.
\begin{figure}[t]
	\centering
	\includegraphics[width=0.16\linewidth]{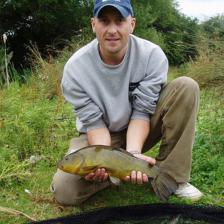}
	\includegraphics[width=0.16\linewidth]{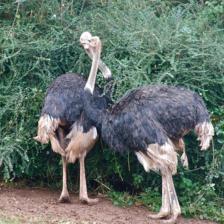}
	\includegraphics[width=0.16\linewidth]{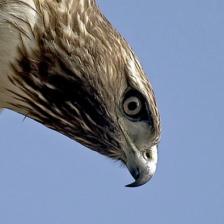}
	\includegraphics[width=0.16\linewidth]{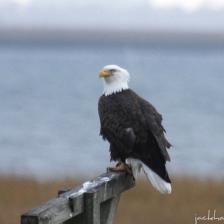}
	\includegraphics[width=0.16\linewidth]{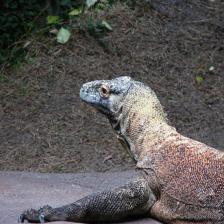}
	\includegraphics[width=0.16\linewidth]{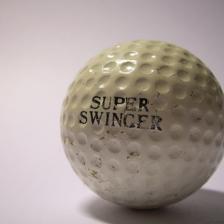}\\
	\includegraphics[width=0.16\linewidth]{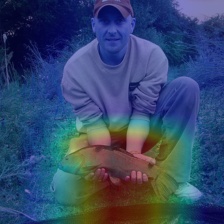}
	\includegraphics[width=0.16\linewidth]{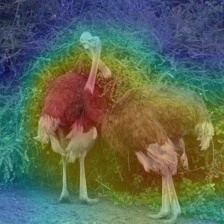}
	\includegraphics[width=0.16\linewidth]{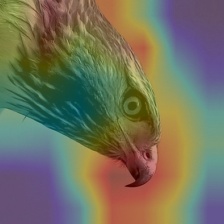}
	\includegraphics[width=0.16\linewidth]{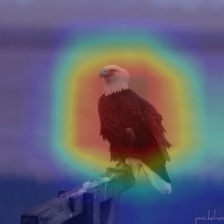}
	\includegraphics[width=0.16\linewidth]{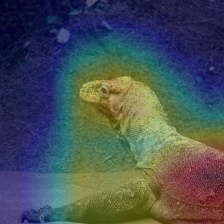}
	\includegraphics[width=0.16\linewidth]{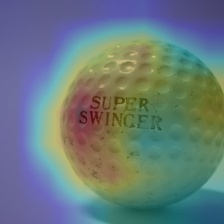}\\
	\includegraphics[width=0.16\linewidth]{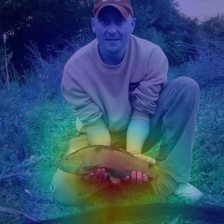}
	\includegraphics[width=0.16\linewidth]{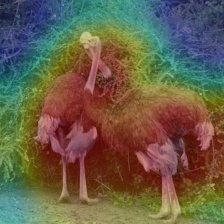}
	\includegraphics[width=0.16\linewidth]{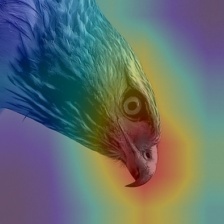}
	\includegraphics[width=0.16\linewidth]{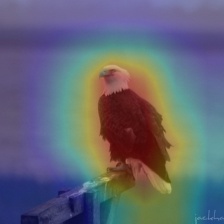}
	\includegraphics[width=0.16\linewidth]{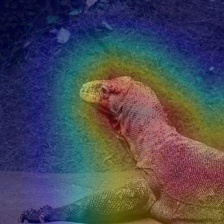}
	\includegraphics[width=0.16\linewidth]{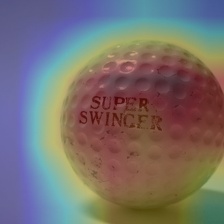}
	\caption{\textbf{Class Activation Maps (CAMs).} The first row shows the original image. The second row shows the images with CAMs for ResNet-50 (baseline model) trained on ImageNet. The third row shows the images with CAMs for ShuffleBlock based ResNet-50 trained on ImageNet.}
	\label{fig:CAM}
\end{figure}
\subsection{Activation Maps}
We use Class Activation Maps (CAM) proposed in \cite{zhou2016learning} to visualize the discriminative regions the model is focusing on to identify the class. We use ResNet-50 for the experiments on ImageNet. We extract class activation maps from ShuffleBlock-based ResNet-50 as well as from its corresponding baseline model. Fig. \ref{fig:CAM} shows the comparison of the activation maps obtained from the baseline ResNet-50 model pretrained on ImageNet with the activation maps obtained from ResNet-50 trained with ShuffleBlock on ImageNet. It can be observed that the ShuffleBlock based ResNet-50 has a better spatial distribution of the activated region in comparison to its corresponding baseline model.

\section{Conclusion}
This paper explores the channel shuffling operation as a regularization technique in  convolutional neural networks. We propose a method called ShuffleBlock and  show that randomly  shuffling  small  patches between channels significantly improves the performance of convolutional neural networks. The patches to be shuffled are picked from the same spatial locations in the feature maps such that a patch when transferred form one channel to another acts as structured noise for the channels. Furthermore, we propose Inverted Step Scheduling that when used with ShuffleBlock improves its performance. We show that ShuffleBlock improves the performance of several baseline networks  on the task of image classification on CIFAR and ImageNet datasets. We provide several ablation studies on selecting various hyperparameters for the ShuffleBlock module. 

%
%
\bibliographystyle{splncs04}
\bibliography{main}
\end{document}